\documentclass[eat,twocolumn]{jmlr}

\usepackage{longtable}

\usepackage{booktabs}
\usepackage[load-configurations=version-1]{siunitx} 

\theorembodyfont{\upshape}
\theoremheaderfont{\scshape}
\theorempostheader{:}
\theoremsep{\newline}

\jmlrvolume{}
\firstpageno{1}

\jmlryear{2022}
\jmlrworkshop{Machine Learning for Health (ML4H) 2022}

\usepackage[font={small}]{caption}
\newcommand{\epoct}{ePOCT\xspace}

\newcommand{\cd}{CDSS\xspace}
\newcommand{\md}{MoDN\xspace}
\newcommand{\state}{\emph{state}\xspace}

\renewcommand{\ss}{\mathbf{s}}
\providecommand{\R}{\mathbb{R}} 
\providecommand{\cE}{\mathcal{E}}
\providecommand{\enc}[1]{\ensuremath{\cE_{q_#1}}}
\providecommand{\cD}{\mathcal{D}}
\providecommand{\cS}{\mathcal{S}}
\usepackage{float}

\title[Modular Clinical Decision Support Networks]{Modular Clinical Decision Support Networks (\md)---Updatable, Interpretable, and Portable Predictions for Evolving Clinical Environments}
\author{\Name{Cécile Trottet} \Email{cecileclaire.trottet@uzh.ch} \\
\Name{Thijs Vogels} \Email{thijs.vogels@epfl.ch} \\
\Name{Martin Jaggi} \Email{martin.jaggi@epfl.ch} \\
\Name{Mary-Anne Hartley} \Email{mary-anne.hartley@epfl.ch} \\
\addr Ecole Polytechnique Fédérale de Lausanne, Lausanne, Switzerland}

\begin{document}

\maketitle

\begin{abstract}
Data-driven Clinical Decision Support Systems (\cd) have the potential to improve and standardise care with personalised probabilistic guidance. However, the size of data required necessitates collaborative learning from analogous \cd's, which are often unsharable or imperfectly interoperable (IIO), meaning their feature sets are not perfectly overlapping. We propose Modular Clinical Decision Support Networks (\md) which allow flexible, privacy-preserving learning across IIO datasets, while providing interpretable, continuous predictive feedback to the clinician.

\md is a novel decision tree composed of feature-specific neural network modules. It creates dynamic personalised representations of patients, and can make multiple predictions of diagnoses, updatable at each step of a consultation. 
The modular design allows it to compartmentalise training updates to specific features and collaboratively learn between IIO datasets without sharing any data.
\end{abstract}

\section{Introduction}
\label{sec:intro}
In this work, we propose the Modular clinical Decision Support Network (\md) to provide dynamic probablistic guidance in decision-tree based consultations. The model is extended during the course of the consultation, combining neural network \textit{modules} specific to each question asked. This results in a dynamic representation of the patient able to predict the probability of various diagnoses at each step of a consultation.

We validate \md on a real world Clinical Decision Support System (\cd)-derived data set of $3000$ pediatric outpatient consultations and show how the feature-wise modular design addresses collaborative learning from imperfectly interoperable (IIO) datasets with systematic missingness, while improving data availability, model fairness and interpretability.
The feature-wise modularisation makes \md interpretable-by-design, whereby it aligns its learning process with the clinician, in step-by-step ``consultation logic''. It can thus provide continuous feedback on each question, allowing the clinician to directly assess the contribution of each feature to the prediction at the time of feature collection.
A major contribution is the possibility to compartmentalise updates to features of interest, thus retaining validity despite the evolving nature or availabilty of features.
\section{Related work}
Most classical machine learning models handle missing features by using imputation, at the risk of injecting bias. MoDN however handles missingness by design, by only applying the encoder modules corresponding to available features. Few other options exist for imputation-free neural nets. For instance, \citet{Sharpe1995NeuralSystems} propose a single neural net for each possible configuration of complete features; an idea that has since been iterated by \citet{Krause2003-ensemble} and \citet{Baron2021-meta}. However, all these approaches suffer at scale, where the number of possible configurations grows exponentially with the number of features, creating an unfeasible computational overhead in high-dimensionality data sets. By comparison, the number of neural networks in our approach is linear in the number of features and targets, and thus scales reasonably well to the high-dimensional space. 

Our work is inspired by the modular neural networks (MNN) \citep{Shukla2010} architecture. In MNNs, neural networks are used as modules and work together to perform the learning task. MNNs are often preferred over monolithic networks to treat high-dimensional input spaces. We are aware of only a few examples of work related to the use of MNNs in the medical field. The only applicable example we found wasby \citet{Pulido2019BloodNetworks} who used modular networks to diagnose hypertension. They use three different modules, each processing a subset of the data related to a specific aspect of the disease. In our model, we used a different module for each distinct feature in the data set, since no subgroups of features are constrained to be always present together. 
\section{Model architecture}
\md comprises three core elements: \emph{encoders}, \emph{decoders}, and the \state as summarised in \autoref{fig:model_architecture}.
\begin{itemize}
    \item The \state, $\ss$, is the vector-representation of a patient. It evolves as more answers are recorded.
    \item \emph{Encoders} are feature-specific and update the \state with the value of a newly collected feature.
    \item \emph{Decoders} are output-specific and extract predictions from the \state at any stage of the consultation.

\end{itemize}

\begin{figure*}[htbp]
    \floatconts
      {fig:model_architecture}
  {\caption{\textbf{The Modular Clinical Decision Support Network (\md)}. The \state is a representation of the patient, which is sequentially modified by a series of inputs. Here, we show in blue \emph{age} and \emph{fever} values as modifying inputs.  Each input has a dedicated encoder which updates the \state. At any point in this process, the clinician can either apply new encoders (to update the \state) or decode information from the \state (to make predictions, in green).}}
  {\includegraphics[scale = 0.6]{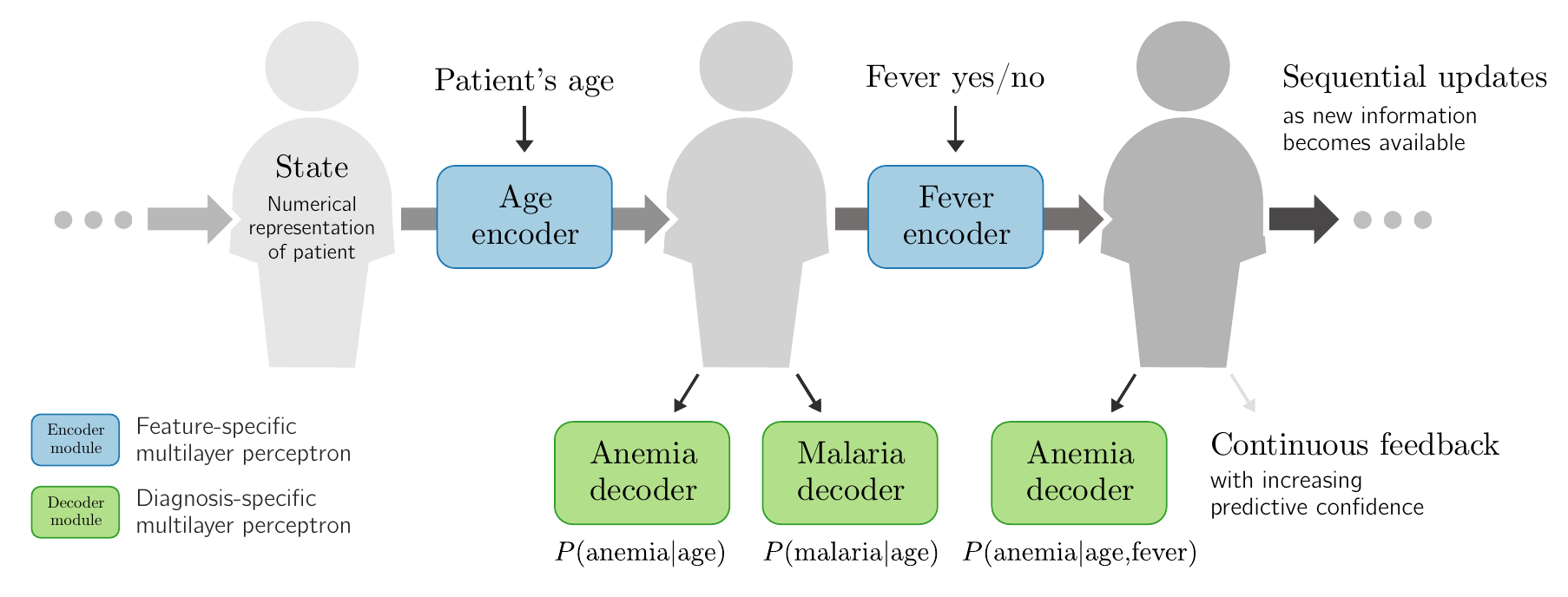}}
    
\end{figure*}
\begin{figure*}[htbp]
\includegraphics[scale = 0.45]{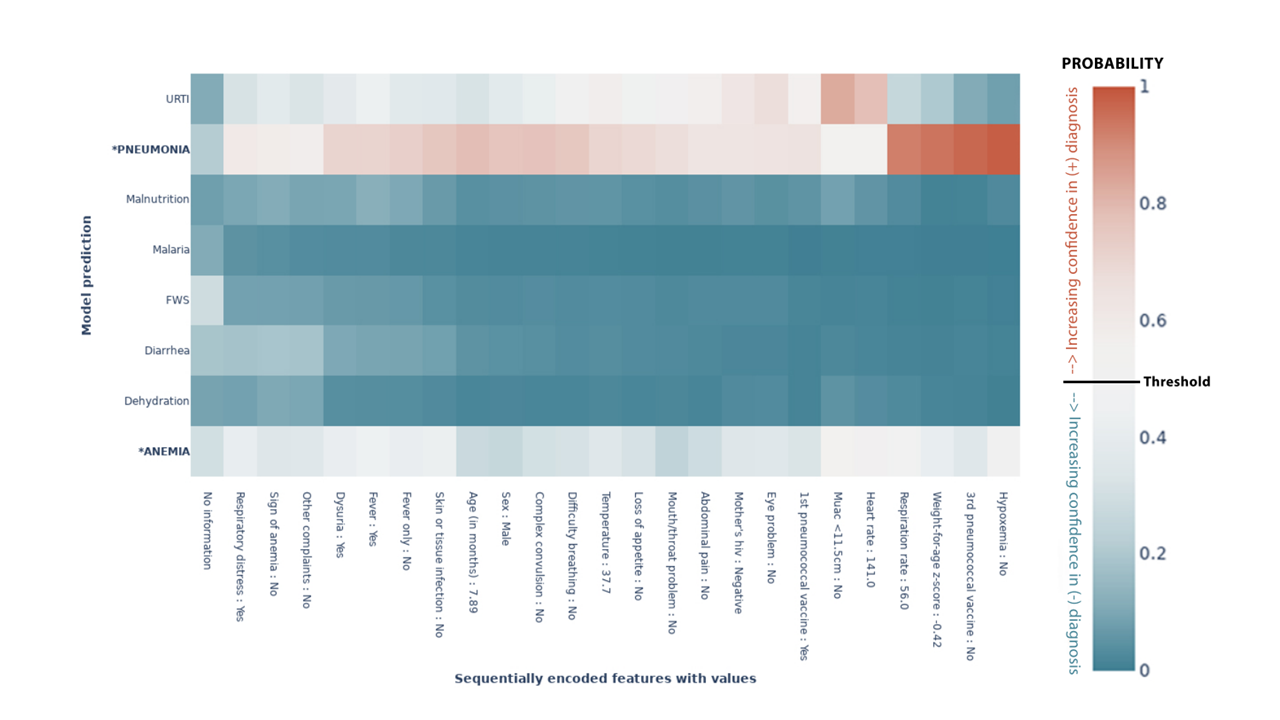}
    \caption{\label{fig:heatmaps}
    \textbf{\md's feature-wise predictive evolution in a patient.} The graph represents a single patient from the test set. The $y-$axis lists the eight possible diagnoses predicted by our model. The true diagnosis of the patient is in bold and marked by an `*'. The $x-$axis is a sequential list of questions asked during the consultation (the response of that specific patient is also listed). The heatmap represents a scale of predictive certainty from red to blue. The patient has true diagnosis of pneumonia and anemia. Here, predictive confidence accumulates slowly throughout the consultation. 
    \textit{\textbf{*}: True diagnosis, URTI: Upper Respiratory Tract Infection, FWS: Fever Without Source, Threshold: probability at which the model categorises the patient with a diagnosis(50\%)}}
    
\end{figure*}

\noindent Encoders and decoders are thus respectively feature- or output-specific multilayer perceptrons (MLP). This modularises both the input space as well as the predictions made in the output space.\\

We consider the consultation data of a patient as an ordered list of (question, answer) pairs, $(q_1, a_1), (q_2, a_2), \ldots, (q_T, a_T)$.
The ordering of questions asked simultaneously is randomized.
As new information is being collected, the \state vector $\ss \in \R^s$ evolves as:
\begin{align}
    \label{equ:state_update}
    \ss_0 &= \cS_0 \in \R^s \text{, a trained constant}, \\
    \ss_t &= \enc{t}(\ss_{t-1}, a_t), \quad \text{for } t = 1\ldots T,
\end{align}

where $\enc{t}: (\R,\R^s) \to \R^s$ is an \emph{encoder} specific to question $q_t$. It is a small MLP with trainable parameters. It updates the state by adding the change to the previous state.

After gathering the first $t$ answers, the probability of a diagnosis $d$ is then predicted as:
\begin{align}
    p_t(d) = \cD_d(\ss_t),
\end{align}
where each \emph{decoder} $\cD_d : \R^s \to [0, 1]$ is also a small MLP with trainable parameters.
\section{Dataset}
We train, test and validate \md on a \cd-derived data set comprising 3,192 pediatric outpatient consultations presenting with acute febrile illness.
The data was collected in Tanzania between 2014 and 2016 as part of a randomised control trial on the effect of \cd on antibiotic use, hereafter referred to as \epoct \citep{KeitelSafetyePOCT}.
The data had over 200 unique feature sets of asked questions (i.e. unique combinations of decision branches in the questionnaire).
\section{IIO experiments}

A common issue when \cd are updated in light of newly available resources (e.g. new questions/tests added to the \cd) is incomplete feature overlap between old and new data sets. To test the capacity of \md in such imperfectly interoperable (IIO) settings, we simulate IIO subsets $A$ and $B$ within the \cd-derived dataset. Performance is then evaluated on an independent test set, in which all of the features are available. Three levels of IIO (from 60---100\%) are simulated between data sets $A$ and $B$  by artificially deleting features in $A$. 

\subsection{New IIO user experiment: Modularised fine-tune}
In this scenario, a clinical site starts using a \cd. It has slightly different resources and is thus IIO compared to more established implementation sites. However, it would still benefit to learn from these sites while ensuring that the unique trends in its smaller, local data set are preserved. 

With \md, we can port the relevant modules pre-trained on the larger more established \textit{source} data set ($A$) to another smaller and IIO \textit{new} data set ($B$) and fine-tune them whilst adding additional modules. \autoref{fig:iioexperiments} shows the performance in macro $F_1$ score for this proposed solution (dark grey, \textit{Modular fine-tuning}) tested in three levels of feature overlap, and compared to three baselines described in \autoref{baselines}.

\subsection{New IIO resource scenario: Modularised update}
In this scenario, a site using a \cd acquires new resources. It would like to update its \cd model with the data collected from this new feature, but it cannot break the validity of the existing predictions that have been approved by the regulatory authority after a costly validation trial. 

\md isolates training updates to newly added features. We use the trained modules from the source data set as a starting point and keep their parameters fixed. We then apply the frozen modules to the local data set and only train the modules corresponding to new features. The performance of this model is shown in light grey in \autoref{fig:iioexperiments}. Similarly to the modularised fine-tuning, we see that modularised update matches the global model where all data is shared. This shows that \md decision rules can be adapted to new features without modifying previously validated predictions of existing features, thus preserving the validity of the tool. 

\subsection{Baselines}
\label{baselines}
We compared \md's performance in these two scenarios to three baselines. 
\textbf{The static model} is where modules trained in $A$ are directly tested in $B$, thus not considering additional IIO features.
\textbf{The local model} is where modules are only trained on the target data set $B$, thus without insights from the larger source data set.
\textbf{The global model} is the ideal, but unlikely, scenario of when all data can be shared between $A$ and $B$ and the modules are trained on the union of data ($A \cup B$).

Performance of upper and lower baselines and \md are compared in \autoref{fig:iioexperiments}. More detail about \md and the baselines in the IIO settings can be found in appendix \autoref{apd:second}. 
\begin{figure}[t]
\floatconts
{fig:iioexperiments}
    {\caption{
       \textbf{ Comparison between ported models and baselines.}  Performance metric is the mean macro F1 scores with 95\% CIs.
        Modularised fine-tuning or updating on additional local features (\textbf{grey}) consistently increases the model's performance
        compared to statically using a source model that only uses shared features (\textbf{teal}).
        The modularised update scenario achieves this without changing the model's behaviour on patients in the source dataset.
        The fine-tuning approaches perform almost as well as the global baseline (\textbf{purple}) that trains on the union of shared data.
        When the percentage of shared features is 80 or 100\%, fine-tuning is significantly better than training only locally on the small `target' dataset (\textbf{green}).
    }}
    {\includegraphics[scale=0.6, trim=6 6 0 0, clip]{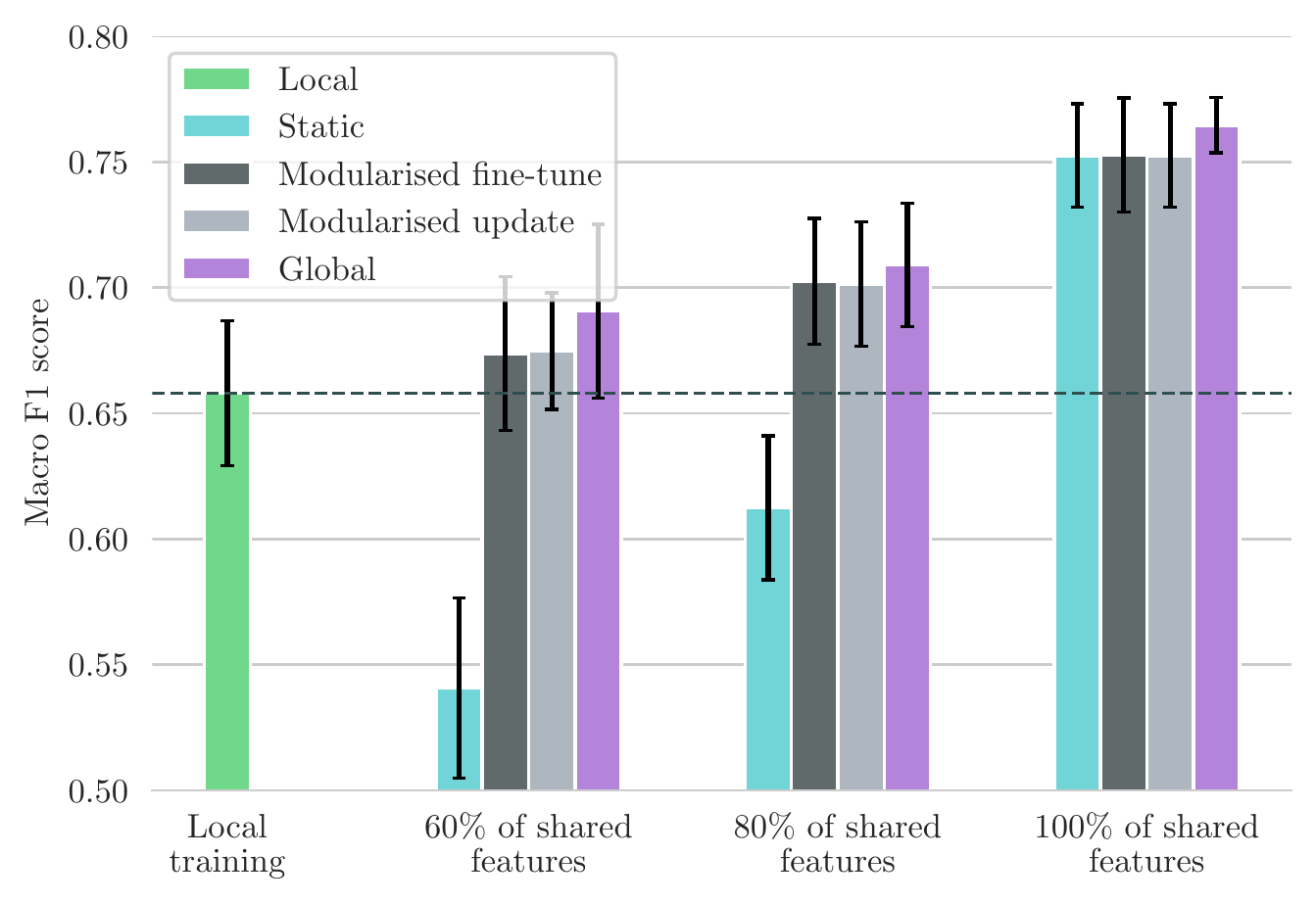}}
\end{figure}

\section{Visualizing diagnostic trajectories}
 One of the assets of \md is that it provides the clinician with feedback at any point in the consultation. For a given patient, the clinician can thus see how the predictions evolve as the features are encoded. The heatmap in \autoref{fig:heatmaps} shows the predictive evolution for a patient of the test set.  We see how the model shifts towards the colour poles (blue and red extremes) as it learns more about the patient and gains confidence. 

These feature-wise predictions thus give the clinician an assessment of the impact of that feature on the prediction.

\section{Discussion}
\md shows the various advantages of modularising neural nets into bite-sized predictions, which not only improves predictive performance on CDSS-derived data (predictive performance is compared to traditional baselines in appendix \autoref{apd:first}) but also allows it to interpretably integrate into the sequential logic of a medical consultation.
The flexible portability of the modules also provides more granular options for building collaborative models, which may address some of the most common issues of model validity when trained and applied to IIO datasets \citep{doi:10.1056/NEJMc2104626} as well as data ownership and privacy.

\acks{The authors thank the patients and caregivers who made the study possible, as well as the clinicians who collected the data on which \md was validated. \citep{KeitelSafetyePOCT}}

\section{Data and code availability}
Anonymized data are publicaly available here: \url{https://zenodo.org/record/400380#.Yug5kuzP00Q}

\noindent The full code are available at the following GitHub repository: \url{https://github.com/epfl-iglobalhealth/MoDN-ML4H}
\clearpage

\bibliography{pmlr-sample}

\appendix
\section{Data availability}
Anonymized data are publicaly available here \url{https://zenodo.org/record/400380#.Yug5kuzP00Q}
\section{MoDN Diagnosis decoding}\label{apd:first}
\begin{figure}[htbp]

    \centering
    \includegraphics[width=0.5\textwidth]{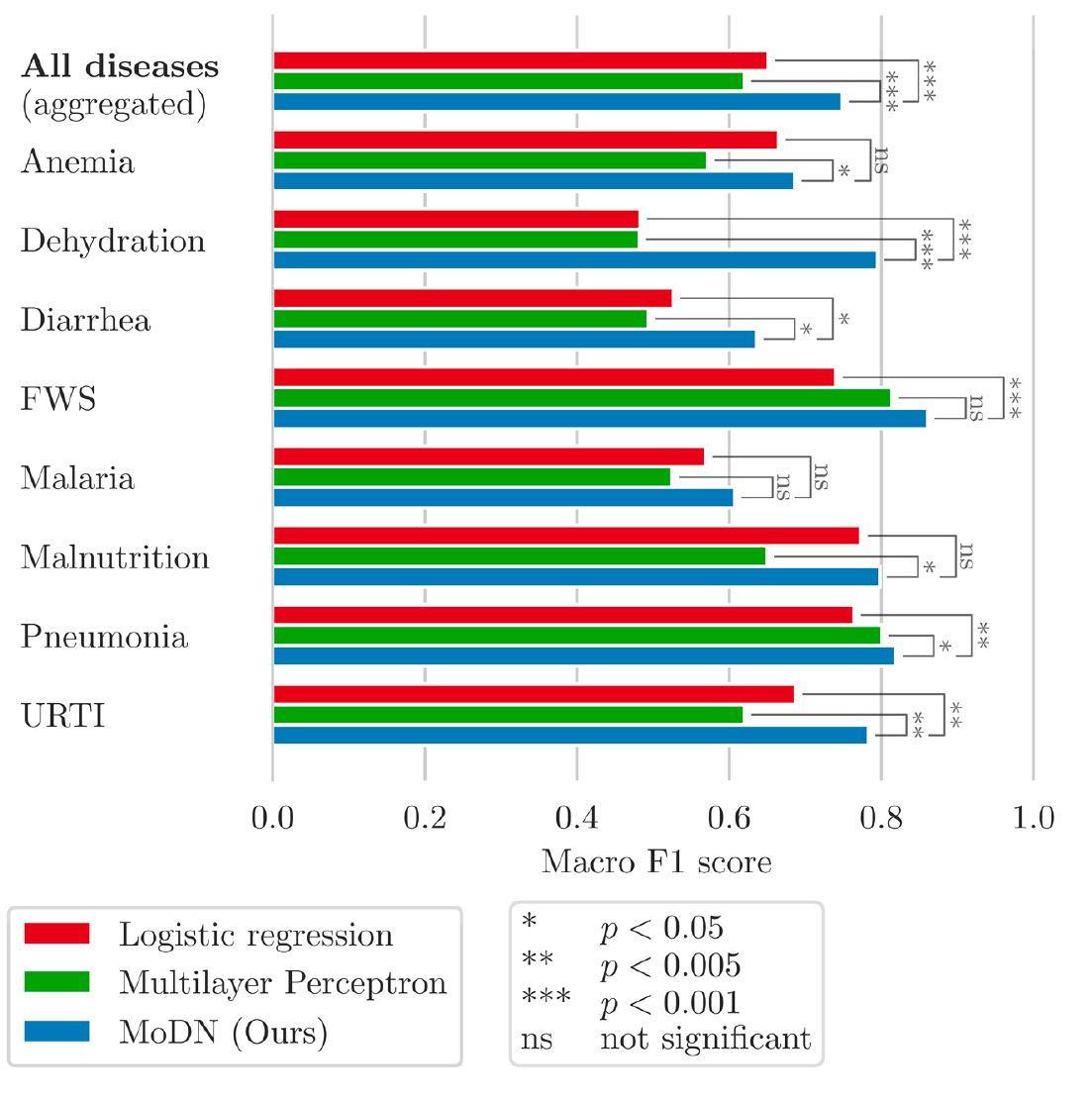}

    \caption{\textbf{MoDN diagnosis decoding performance} Mean of the $5\times 2$ cross-validated macro F1 scores for the diagnosis prediction on the test sets. Furthermore, \md significantly beats at least one of the baselines for each of the individual diagnoses except for malaria. }
    \label{fig:results_basic}
\end{figure}

The predictive performance of \md was compared to the best logistic regression and MLP algorithms for each target diagnosis. \autoref{fig:results_basic} shows the macro $F_1$ scores (unweighted average of $F_1$ scores for the presence and absence of the disease). An overall performance is computed as the average over all diagnoses. Paired \emph{t}-tests show that \md significantly outperforms the baselines for all binary classifications as well as for the overall diagnosis prediction. Malaria is an exception, where \md and baseline models have equivalent performance.

\section{IIO models and baselines}\label{apd:second}
\begin{figure*}[htbp]
    \centering
    \includegraphics[width=1\textwidth]{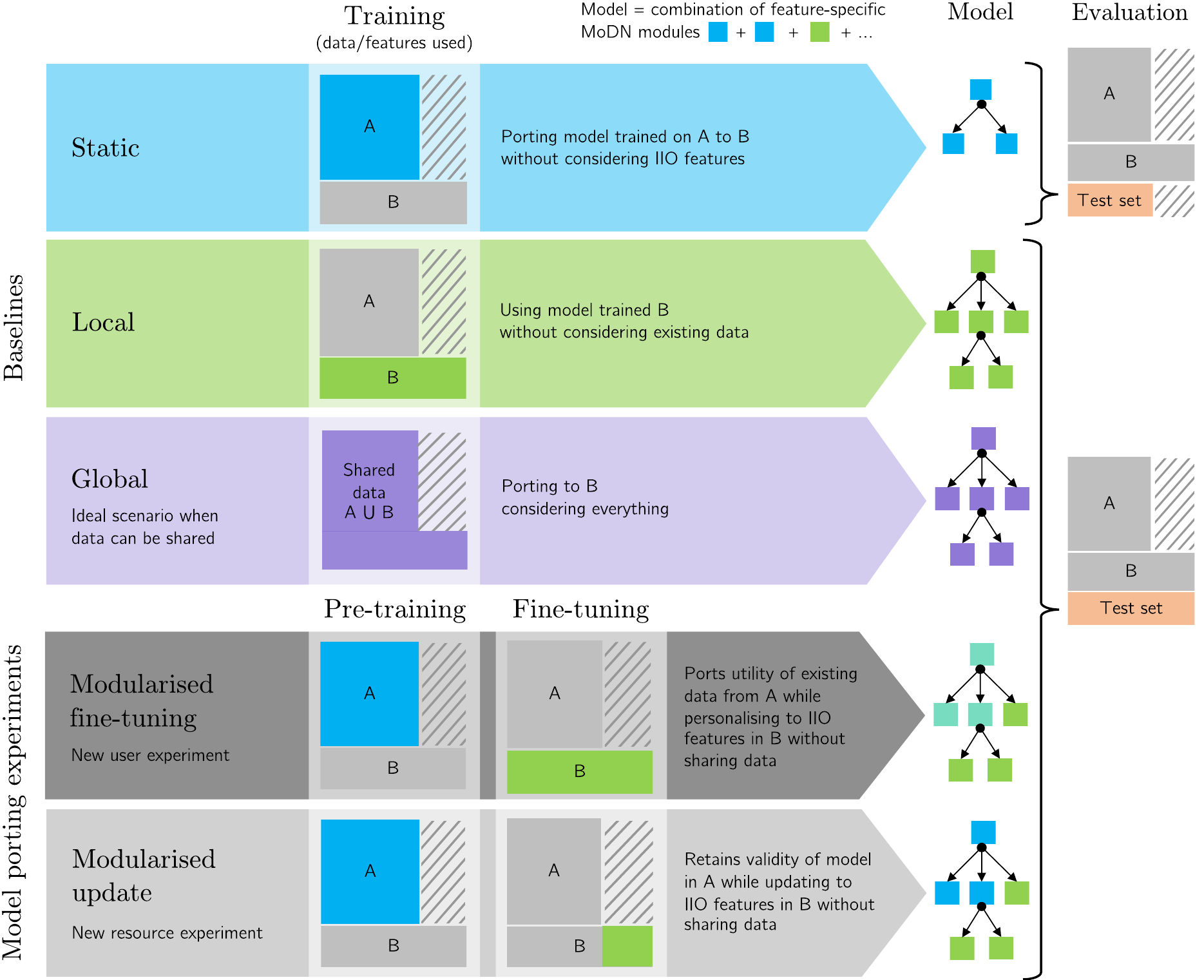}
        \caption{ 
        \textbf{Experimental set up for porting MoDN modules in IIO settings.} \md is tested in two "model porting experiments" (grey), where modules are ported from a larger \textit{source} data set ($\mathbf{A}$) for fine-tuning or updating on a smaller, imperfectly interoperable \textit{target} data set ($\mathbf{B}$). The two experiments represent either a scenario where a new user with different resources starts using a \cd or where an existing user gains new resources and would like to merge training.
        Three baselines are proposed. \\
        \textbf{Static} (blue) where modules trained in $\mathbf{A}$ are directly tested in $\mathbf{B}$, thus not considering additional IIO features. 
        \textbf{Local} (green) where modules are only trained on the target data set $\mathbf{B}$, thus without insights from the larger source data set.
        \textbf{Global} (purple) is the \textit{ideal} but unlikely, scenario of when all data can be shared between $\mathbf{A}$ and $\mathbf{B}$ and the modules are trained on the union of data ($\mathbf{A \cup B}$).
        The \textbf{modularised fine-tuning} experiment, pre-trains on $\mathbf{A}$ and then fine-tunes all modules (for all features) on $\mathbf{B}$ (thus personalising the modules trained on $\mathbf{A}$).
        The \textbf{modularised update} experiment, pre-trains the blue modules on $\mathbf{A}$ and then adds modules specific to the new IIO features (in green) which have been independently trained on $\mathbf{B}$ (thus preserving the validity of the modules trained on $\mathbf{A}$). The colors of the \md modules illustrate their training on distinct data sets and their potential re-combination in the porting experiments. In particular, the modules trained on $\mathbf{A}$ (blue) and fine-tuned on $\mathbf{B}$ (green) are thus depicted in teal.
        }
    \label{fig:baselines}
\end{figure*}

\autoref{fig:baselines} provides a detailed overview of the characteristics and differences between \md in the two IIO settings and the baselines.
\end{document}